# Neuro-Symbolic Closed-Loop Control of Laser Powder Bed Fusion with an In-Loop Ontology

Gisuk Hong[a], Jaebong Cho[a], Hyunbo Cho[a, *]

[a]*Department of Industrial and Management Engineering, Pohang University of Science and Technology, 77 Cheongam-ro, Nam-gu, Pohang, 37673, South Korea*

Corresponding author. Tel.: +82-10-3503-2204. *E-mail address:* hcho@postech.ac.kr

**Abstract**

A geometry-conditioned, neuro-symbolic closed-loop architecture is proposed for laser powder bed fusion, in which a standards-aligned ontology operates inside the control loop and couples symbolic reasoning with statistical learning to set the targets of a constraint-aware predictive controller. The ontology links the process objectives and constraints to the signals a controller can observe, and a description-logic reasoner converts them into the references and bounds enforced on each scan. The demonstrated case is overhang dross: a quality limit on the melt pool depth, which governs quality yet cannot be measured during the build, is mapped through a geometry- and power-dependent depth-to-width ratio onto a bound on the observable width, with the ratio and its calibrated uncertainty supplied by a Gaussian process. The reasoner classifies each upcoming feature and selects the active constraints—adding a lack-of-fusion floor at overhangs, a monotone guard beyond the calibrated range, and an energy-density cap where a process window is declared—while running only on changes of geometric context and otherwise leaving a single small quadratic program on the per-scan path. In an Eagar–Tsai surrogate calibrated to the NIST AM-Bench benchmark for IN625, the architecture eliminates the dross produced by a geometry-blind controller, holds dross at zero with only a small residual lack of fusion under dual scoring, degrades gracefully under deliberate plant mismatch, and retargets to new alloys and constraints by editing ontology data rather than code. The results establish architectural feasibility; experimental calibration of the ratio is the principal next step.

# 1. Introduction

Additive manufacturing (AM) builds a part directly from a digital model by adding material layer upon layer, which makes geometries possible that machining and casting cannot produce. Laser powder bed fusion (LPBF) is the most widely used metal AM process (DebRoy et al., 2018). A laser scans a thin bed of metal powder and melts it along a programmed path, and the small volume of molten metal that forms under the laser, the melt pool, fuses each layer to the one below. The quality of the finished part is set locally by this melt pool, and keeping it within the right size is the central process-control problem in LPBF.

A given set of laser settings does not produce the same melt pool everywhere in a part. Because the surrounding geometry governs how heat leaves the pool, the same command yields a different pool size as the local shape and scan pattern change, for example near thin walls, sharp corners, or downward-facing surfaces. A particularly damaging case is the overhang, the downward-facing region of a feature that is built over loose powder rather than over solid metal. The powder beneath an overhang carries heat away much more poorly than solid metal, so heat cannot escape downward and the pool grows deeper than intended; once it is too deep it melts into the loose powder underneath and solidifies as a rough projection called dross (Yeung et al., 2019; Viale et al., 2022). The defect is caused by the geometry, not by an error in the nominal settings.

Work on this geometry dependence falls into two groups that each address only one side of the problem. Closed-loop controllers sense the melt pool during the build and adjust the laser power in real time (Renken et al., 2019; Papatheodorou et al., 2025), but most regulate a single melt pool signal without conditioning it on geometry, so they react to a shape change only after the pool has changed; the few that do account for geometry either treat the quality-determining depth as a directly available output or lock the geometry dependence inside a learned policy (Liao-McPherson et al., 2024; Park et al., 2024). Geometry-aware feedforward schemes take the opposite approach, precomputing the power from the part geometry before the build (Yeung et al., 2019; Druzgalski et al., 2020), but because they never read the pool they cannot reject the run-to-run variation that sensing handles. No existing method is at once geometry-aware, driven by in-process sensing, and reliant only on a signal that can be measured during the build. Section 2.1 reviews both groups.

Symbolic knowledge offers a way to make this geometry dependence explicit. In additive manufacturing it has been used to formalize processes, parameters, and defects for monitoring, design, and quality management (Sanfilippo et al., 2019; Hasan et al., 2023), but in each case outside the real-time control loop. The complementary idea of placing symbolic reasoning alongside learned or numerical models inside a control loop has shown value in robotics and in process and mechanical control (Sun and Shoukry, 2024; Alhajeri et al., 2022), yet it is largely absent from additive manufacturing. Section 2.2 reviews this work.

This paper addresses that gap by placing geometric knowledge inside the LPBF control loop. We propose a geometry-conditioned, neuro-symbolic closed-loop architecture in which an ontology converts the local geometry and material into the references and constraints of a constraint-aware model predictive controller (MPC) before each scan. This unites the two groups the literature has kept apart: the architecture is geometry-aware, like the feedforward methods, and driven by in-process sensing, like the closed-loop methods, with the geometric knowledge held in an explicit, reusable symbolic layer rather than precomputed offline or buried in a learned policy.

The contribution is this architecture rather than any single mapping inside it: the ontology connects the controller's objectives and constraints, which may concern a quantity the controller cannot observe directly, to the signals it can, and a reasoner maintains that connection as the geometric context changes. The case we evaluate rests on a basic asymmetry of in-process LPBF sensing, that the melt pool depth governs quality but cannot be measured during the build while the width can be observed from the surface; the ontology bridges it by mapping the observed width to an admissible width that acts as a virtual depth sensor, and the controller drives the width to that value. Two features separate this symbolic layer from a typed configuration file: a description-logic reasoner decides, in the loop, which constraints are active for each geometric context, and new constraints or materials are added as data rather than code; the statistical component, in turn, reports a calibrated uncertainty that sizes the controller's safety margin.

The remainder of the paper is organized as follows. Section 2 reviews related work on overhang defects, on in-process and feedforward control, and on knowledge-based methods. Section 3 describes the control architecture in general terms. Section 4 instantiates it in the calibrated surrogate and reports the results.

Section 5 concludes.

# 2. Related work

## 2.1 Sensing and control of the geometry-dependent melt pool

The melt pool is the natural target for control in LPBF because it mediates between the laser settings and the final quality, but it responds to more than those settings. Since the surrounding material sets the rate at which heat leaves the pool, the same command produces a different pool wherever the local shape changes, and the overhang is the limiting case: the loose powder beneath a downward-facing surface conducts heat away far less effectively than solid metal, so the pool deepens and breaks into the powder as dross. Yeung et al. (2019) describe this mechanism and tie it directly to overhang defects, and Zhang et al. (2019) give it a quantitative basis by measuring the conductivity of the powder at roughly 4 to 7 percent of the solid, while Viale et al. (2022) confirm in a review of downskin processing that the remedy is to lower the energy input where heat cannot escape. The shared message of this body of work is that the depth response of the pool, and therefore the tendency to form dross, is a function of the geometry rather than a fixed property of the material. It also exposes the difficulty any controller faces: the defect is set by a subsurface depth that cannot be measured during the build.

One response is to close a loop around the pool during the build. Renken et al. (2019) stabilize the melt pool temperature by modulating the laser power in real time, combining feedback with a feedforward term, and report how tight the timing budget is at LPBF scan speeds; Wang and Chen (2021) drive a feedback loop from a finite-element model so that a melt pool output, such as width, updates the power command. These controllers regulate whatever signal the sensor provides, but they do not condition the target on the geometry, so a shape-induced change is corrected only after it has already disturbed the pool, and a recent review confirms that geometry-aware closed-loop control remains an open problem in this space (Papatheodorou et al., 2025). Two recent controllers come closer and are worth separating from the rest, because the present work is defined partly by how it differs from them. Liao-McPherson et al. (2024) regulate melt pool depth on overhang geometries with a model-based, layer-to-layer law; their target is the quality-determining dimension itself, but they treat that depth as an available output and act between layers, whereas the controller proposed here treats depth as unmeasurable, infers it from the observed width, and acts from track to track within a layer. Park et al. (2024) reach geometry awareness from the opposite direction, learning a reinforcement-learning policy that maps in-situ measurements and geometric information to the laser power; this couples sensing and geometry inside the loop, but the dependence is locked inside a trained policy, whereas the present work keeps it in an explicit, inspectable form that can be examined and edited without retraining.

A second response keeps the geometry but gives up the sensor, precomputing the laser power from the part geometry before the build. Yeung et al. (2019), in the feedforward half of their work, scale the power by a geometric conductance factor read from the local proportion of solid and powder; Wang et al. (2020) invert a control-oriented melt pool model to obtain a power trajectory that avoids over-melting and keyholing; Druzgalski et al. (2020) assign parameters to individual scan vectors from a simulation library at part scale; and Reiff et al. (2021) adapt power, speed, and scan strategy from a thermal model evaluated during the build. These methods establish that conditioning the settings on geometry is both feasible and effective, which is the half of the problem the closed-loop controllers leave out, but because they never read the pool they cannot reject the run-to-run variation, such as drifting absorptivity or delivered power, that closed-loop sensing exists to handle. The two responses therefore leave a gap between them: one corrects disturbances but is blind to geometry, the other anticipates geometry but is blind to disturbances, and none of them produces a controller that is geometry-aware, driven by in-process feedback, and reliant only on a signal that can actually be measured. Closing that gap is the aim of the architecture in Section 3.

### 2.2 Knowledge representation and neuro-symbolic control

The geometry dependence above is exactly the kind of structured, reusable engineering knowledge that a symbolic representation is meant to capture, and a substantial line of work has built such representations for additive manufacturing. Sanfilippo et al. (2019) propose a vendor-independent ontology that organizes AM data and reasons over it to validate that a configuration is consistent, for instance that a machine and a material are compatible. Ko et al. (2021) combine machine learning with a knowledge graph to derive design rules, and parameterize an overhang feature by its downskin angle, which is close in spirit to the geometric conditioning used here, though their rules are applied when the part is designed. Yang et al. (2020) place ontology analytics inside a quality-management framework aimed at post-build inspection and root-cause analysis, and Hasan et al. (2023) build an ontology of LPBF defects and their causes and use a reasoner to infer new causal links over it. These works show that AM knowledge, including knowledge about overhangs and defects, can be formalized and reasoned over, but they apply it before the build, for design validation, or after it, for diagnosis and quality analysis. A survey of AM monitoring, diagnosis, and control makes the boundary explicit by treating defect inspection and real-time control as separate activities (Fang et al., 2022). The knowledge that could tell a controller how to respond to an overhang therefore exists, but it has not been placed where a controller could use it during the build.

Bringing symbolic knowledge into the loop in this way is the defining feature of neuro-symbolic control, which has matured in fields adjacent to AM. In robotics, Sun and Shoukry (2024) embed a symbolic task model in the training of a neural planner so that the learned controller inherits the interpretability and correctness guarantees of the symbolic part, and Wang et al. (2024) let symbolic reasoning guide neural learning in a framework whose tasks include optimal control. In process and mechanical control, physics-informed predictive controllers carry first-principles structure inside the model that an MPC optimizes, for chemical reactors (Alhajeri et al., 2022; Zarzycki and Ławryńczuk, 2023) and for manipulators (Nicodemus et al., 2022), and broader reviews report that adding explicit knowledge to a control loop improves data efficiency, generalization, and interpretability (Ejalonibu, 2025; Hamilton and Ali, 2026). The lesson from these domains is that a symbolic layer inside the loop can supply structure and transparency that a purely learned controller does not, yet the approach has scarcely been applied to AM, and where symbolic knowledge has been used for AM, it has stayed outside the loop. Placing it inside the loop is the opening that the architecture in Section 3 addresses.

## 3. Method

### 3.1 Overview of the control architecture

The control objective considered in this work is to keep the melt pool within limits that depend on the local geometry, using only signals that are available while the part is being built.

The difficulty that motivates the architecture is that the defects of interest are caused by geometry, while a controller that observes only the melt pool cannot tell a geometric cause from an ordinary process disturbance. A pool that grows because the laser power drifted and a pool that grows because the scan has reached an overhang produce a similar signal, yet they call for different responses. The architecture proposed here places a knowledge layer between the sensing and the control of the machine, so that the reference and bound the controller uses are conditioned on the local geometry before each scan is executed. Stated generally, this layer expresses each control objective or constraint, including one that concerns a quantity the controller cannot sense directly, as a bound on a quantity it can, and the depth-from-width mapping developed in Section 3.2 is the instance of this principle used throughout the paper.

Figure 1 shows the flow of information for a single scan. The diagram separates an offline path, which grounds and assembles the knowledge, from an online loop that runs in real time during the build, and the two meet at the control-prior bus, which carries the per-scan reference and bound from the knowledge layer to the controller. The architecture has four components. The plant is the LPBF process, represented here by a process model that maps the commanded laser power to a melt pool whose width is observed and whose depth is hidden and determines quality. A grounding step reads the geometry and material of the upcoming scan from the build plan and assembles them into the context on which the knowledge layer acts. The knowledge layer stores the calibration data and the relations that connect geometry to admissible melt pool conditions, and a statistical component supplies the coefficients of those relations that vary with context. The predictive controller receives a width reference and a width bound from the bus, computes the command, and applies it before the scan begins, after which the observed width is fed back and the cycle repeats for the next track. Two further elements stand outside the loop and support it without taking part in the control. A provenance log records the source of each calibration coefficient, and a defect label, defined as the true depth exceeding its geometry-dependent limit, is recorded for evaluation only, because it is hidden from the controller during the build. The components themselves, namely the predictive controller, the statistical model, and the ontology, are standard. The contribution is their arrangement into a closed loop in which symbolic knowledge sets the control targets, and the central element that makes this possible, the depth-from-width mapping, is described in Section 3.2.

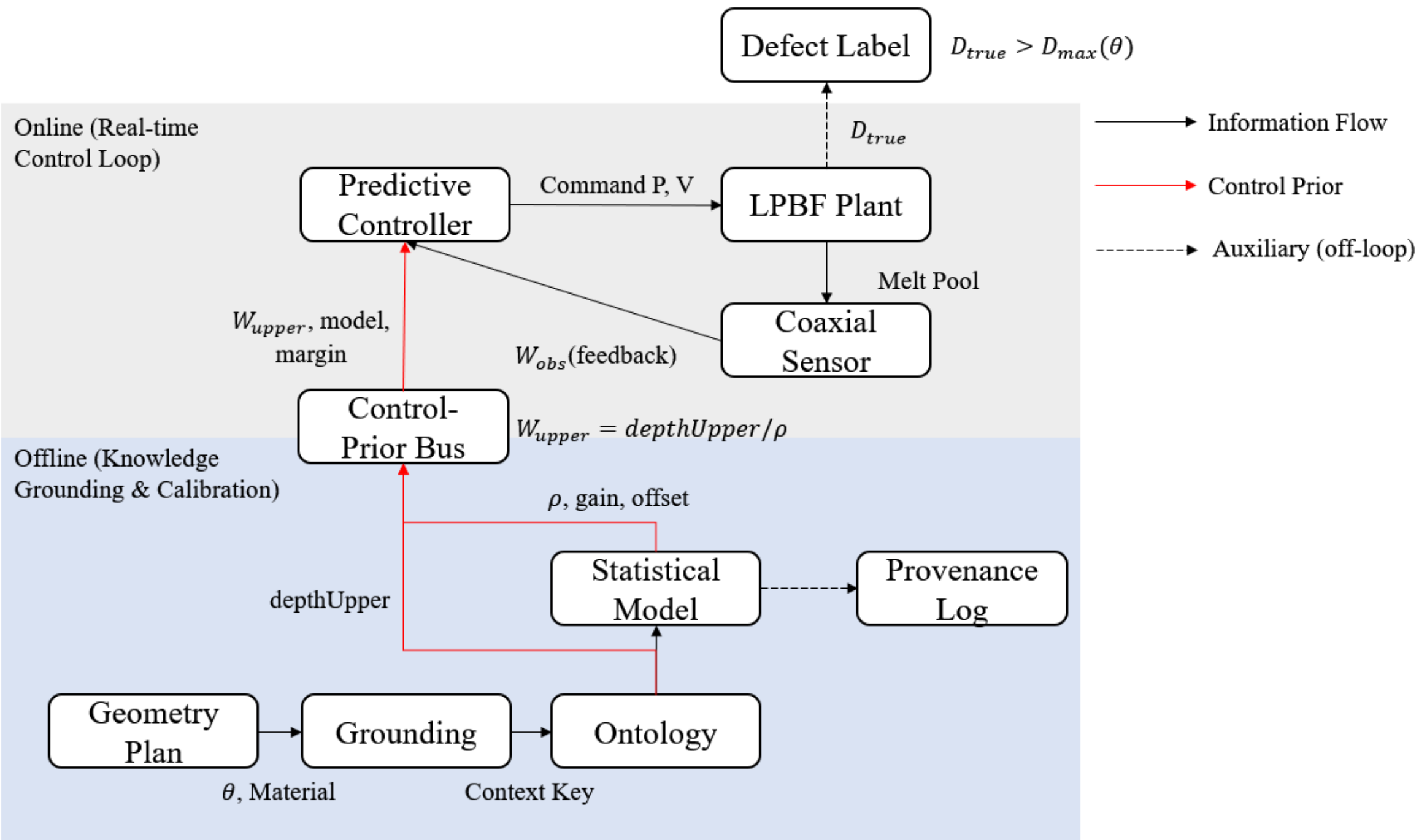

**Figure 1.** Flow of information for a single scan.

## 3.2 Process abstraction and the geometry-dependent depth-to-width ratio

The architecture is built on an abstraction of the melt pool that uses two of its dimensions. The width $W$ is the lateral extent of the pool at the surface, which can be estimated during the build from a coaxial image of the pool. The depth $D$ is the extent of the pool below the surface. Depth is the dimension that governs quality, because a pool that is too shallow leaves a lack-of-fusion void, while one that is too deep beneath an overhang forms dross. Depth cannot be measured during the build, since it is a subsurface quantity, whereas width is accessible to surface imaging. Both dimensions are on the order of tens to low hundreds of micrometers for the conditions considered here.

Geometry enters the abstraction through heat conduction. The downward heat path under an overhang is degraded far more than the lateral one, because the material below is loose powder while the sides are consolidated tracks, so the pool deepens more than it widens as the overhang steepens. The consequence is that the depth-to-width ratio $\rho(\theta) = D/W$ is not a constant of the material but a function of the local overhang angle $\theta$. It increases as the overhang steepens. The ratio depends most strongly on the geometry, but it also varies with the laser power across the operating range, as quantified in Section 4.3. Rather than assume a single value, the knowledge layer holds it as a function of geometry and power and supplies its value at the power the controller is about to command, computed from $D/W$ at that operating point.

This dependence is the reason the knowledge layer is needed: the controller must convert a limit on the unmeasurable depth into a limit on the observable width, and the conversion uses the ratio. Writing $D_{lim}(\theta)$ for the largest depth that does not produce dross at angle $\theta$, the corresponding width bound is $W_{bound}(\theta) = D_{lim}(\theta)/\rho(\theta)$.

A controller that assumes a single value of $\rho$ computes a width bound that is correct only at the angle where that value holds. For an overhang the true ratio is larger than the flat-region value, so the assumed bound is too high, the controller permits a wider pool than intended, and the resulting depth exceeds the limit. We refer to the mapping from the observed width and the local context to the admissible width as a context-conditioned virtual depth sensor. It is the LPBF instance of the general link of Section 3.1, an objective stated on an unobservable variable expressed as a bound on an observable one, and it is the means by which geometric knowledge enters the control law.

## 3.3 Knowledge layer and inference

The knowledge layer is an ontology that reuses established vocabularies by importing them directly, rather than by renaming their concepts. The Basic Formal Ontology (Arp et al., 2015) supplies the upper categories that separate processes from the material entities they act on. The QUDT vocabulary, for Quantities, Units, Dimensions and Types (QUDT, 2021), supplies the description of process parameters as physical quantities. The SOSA vocabulary, for Sensor, Observation, Sample, and Actuator (Haller et al., 2018), supplies the sensing concepts, so that the observable melt pool quantities and the sensor that measures them are described in standard terms. The defect modes are placed under the geometric-flaw class of MASON, a published manufacturing-domain ontology (Lemaignan et al., 2006), so that dross and the other modes are expressed as a recognized kind of manufacturing flaw. Each domain class is asserted as a subclass of the corresponding external class through a genuine import of the source ontology, which keeps the defect and sensing descriptions consistent with established usage and limits the part of the model that must be defined anew.

The layer differs from a fixed rule base in the way it treats numerical coefficients. The form of each relation is separated from the constants that appear in it, and the constants are stored as data with a record of their origin. The relation that gives the admissible depth as a function of overhang angle is stored as a structure, while its coefficients are held as attributes of a calibration entity and are taken from independent process-map literature. The width response to power and the depth-to-width ratio are held as attributes of a second calibration entity, together with the estimated uncertainty of the ratio, and are produced by the statistical component described below. With this organization, revising a coefficient is an edit to data rather than a change to logic, every coefficient carries its provenance, and the same controller can be retargeted to a new material by replacing data.

The relations in the layer are of two kinds, and the distinction matters for the role of inference. The numerical relation that converts an admissible depth into a width bound is evaluated arithmetically at every scan, because the controller needs its value in the loop. The qualitative relations, such as the classification of a feature as prone to dross, are expressed as logical definitions and are used in the loop, through a description-logic reasoner, to select the constraints that are active in each context, as shown in Section 4.4. The arithmetic and the logical parts read the same stored coefficients, so the two views remain consistent.

The statistical component that supplies the learned coefficients is a Gaussian process. The choice follows from two requirements of the framework. The controller needs a measure of confidence in each inferred coefficient and not only a point value, because it folds that confidence into the safety margin and uses it to decide when to fall back on the symbolic guard. The margin below the depth limit is set as a fixed minimum plus a multiple of the predictive standard deviation of the inferred ratio, so the controller automatically keeps a wider margin where the calibration is sparse, such as at steep overhang angles far from the training data, and a tighter one where the ratio is well determined; a Gaussian process supplies exactly this calibrated standard deviation together with the mean. In addition, the calibration data cover only a small number of geometries and materials, and a Gaussian process interpolates smoothly and efficiently from few samples. The inputs are the overhang angle and the material properties that define the context, and the outputs are the gain and offset of the width model and the ratio $\rho$. Because the model returns a value for any context in its input space, the controller can be applied at overhang angles and materials that were not part of the calibration set. An underestimate of $\rho$ is unsafe, because it inflates the width bound and admits dross, so the inference is constrained by a rule taken from the ontology. The ratio is known to increase as the overhang angle decreases, and this monotonic relation is imposed when the controller is asked to operate below the range covered by training. Writing $\theta_{lo}$ for the lowest angle in the training set, the guarded estimate is

$$\rho(\theta) = \max\{\rho_{GP}(\theta), \rho_{lin}(\theta), \rho_{GP}(\theta_{lo})\} \; for \; \theta < \theta_{lo} \tag{1}$$

where $\rho_{GP}$ is the Gaussian-process mean and $\rho_{lin}$ is a monotone linear extrapolation from the lower edge of the training range. The guard does not improve accuracy when the statistical model already extrapolates well, and in that case, it leaves the estimate almost unchanged. Its role is to provide a lower bound on $\rho$ that is the conservative choice for the reason just given: because too small a ratio is the unsafe error, taking the larger of the candidate values raises the assumed ratio, which lowers the width bound and shrinks the commanded pool, biasing any remaining error toward the safe, shallower side rather than toward dross. This is a one-sided safety bias and not a proof that the estimate never falls below the true ratio, which holds only where the linear extrapolation itself lower-bounds it. Figure 2 shows the class structure and the alignment to the reused vocabularies.

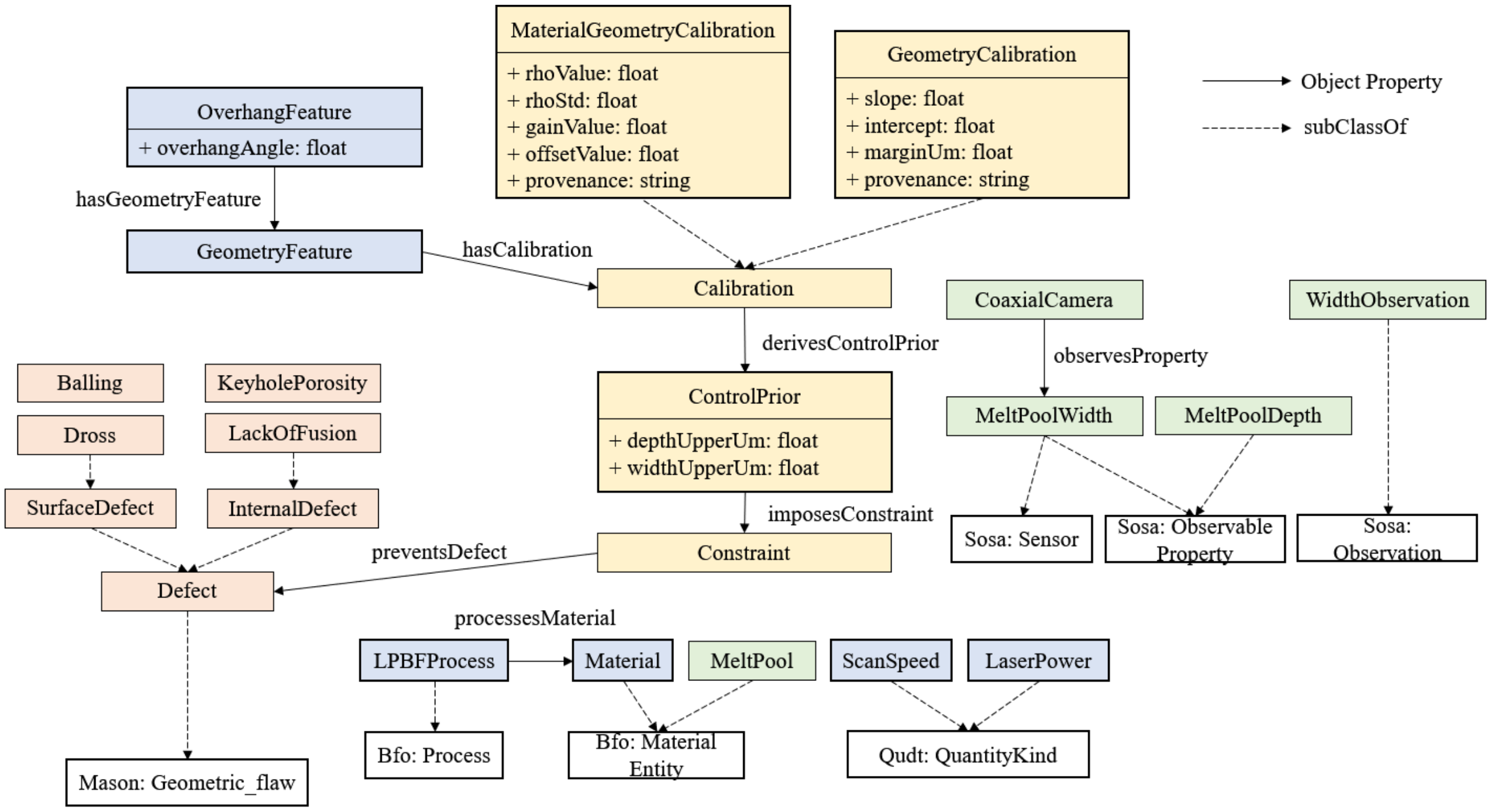

**Figure 2.** Class structure of the ontology and its alignment to the reused vocabularies.

## 3.4 Predictive controller

The controller is a model predictive controller. This form is used because the safety requirement is a constraint on the width rather than a set-point to be tracked, and a predictive controller enforces such a constraint explicitly where a fixed-gain feedback law does not. A predictive controller also looks ahead over a short horizon, which lets the geometry-dependent references and bounds be scheduled along the scan path and applied before a change in geometry is reached, and it accommodates an internal model whose parameters change with context. The controller regulates the observed width. Its internal model relates the width produced by a scan to the commanded power $P$ through a gain and an offset that depend on the context $c$, together with an additive term $e$ that represents the unmodeled disturbance,

$$W = g(c)P + b(c) + e \tag{2}$$

The model is linear in the command, and its parameters $g$ and $b$ vary with the context, so it is a linear parameter-varying model. The knowledge layer supplies $g$ and $b$ for the current context. The model is kept low-order on purpose, because a low-order model keeps the per-scan optimization inexpensive and, together with the disturbance term and the width feedback, is sufficient for control. It does not reproduce the plant exactly, and the term $e$ absorbs the residual mismatch and any slow drift. It is updated after each scan from the measured width by

$$e \leftarrow e + L[W_{obs} - (g(c)P + b(c) + e)] \tag{3}$$

with a fixed gain $L$. This update gives the controller the property of being offset-free, meaning that it removes steady-state tracking error when the disturbance is constant or slowly varying. This property is used in Section 4.2, where heat that accumulates from layer to layer appears to the controller as a slow drift.

Before each scan the controller receives from the knowledge layer the width bound $W_{bound}$defined in Section 3.2 and a width reference $W_{ref}$. The reference is set to the width bound, so that the pool is made as large as the depth limit allows and fusion of the layer below is maintained, and the bound then prevents the depth from exceeding the limit. Over a horizon of $H$ scans the controller solves the quadratic program

$$\begin{gathered} \min\ \textstyle\sum_k \left(W_k - W_{ref}\right)^2 \ +\ \lambda \sum_k (\Delta P_k)^2 \ +\ \mu \sum_k {s_k}^2 \\ \text{subject to}\quad P_{lo} \le P_k \le P_{hi}, \qquad |\Delta P_k| \le \Delta P_{max}, \qquad W_k \le W_{bound} + s_k, \qquad s_k \ge 0 \end{gathered} \tag{4}$$

in which $W_k$ is the predicted width at step $k$ from the model above, $\Delta P_k$ is the change in command between steps, and $s_k$ is a slack variable that keeps the program feasible. The power bounds are not the limits of the laser but a working range qualified for the material, so the flat region operates near the upper bound while an overhang requires a reduction. The rate bound on $\Delta P_k$ represents the fact that a real command is changed gradually between adjacent tracks rather than in a single step. Because the command cannot jump, the look-ahead is what allows the controller to begin lowering the power before an overhang is reached. The first term tracks the reference, the move-penalty term and the rate bound together keep the command smooth, and the slack term with a large weight $\mu$ keeps the width bound respected whenever it is feasible. The controller applies the first command of the solution, observes the width, updates $e$, and repeats at the next track.

The width bound is one of possibly several constraints that the knowledge layer declares. A constraint is stored as data that names a melt pool quantity and an upper value for it, and the controller converts each declared constraint into a bound on the command for the current context and enforces the most restrictive one. A constraint added in this way changes the command without any change to the control law, a property used in Section 4.4 to add a limit on the energy input.

# 4. Case study

The following instantiates the architecture in a calibrated LPBF surrogate and evaluates it on one case study, overhang dross, in which the unobservable objective is the melt pool depth and the observable signal is the width. The case study is the evaluation of this paper; the architecture it exercises, an ontology that connects objectives and constraints to observable signals and a reasoner that selects the active constraints in the loop, is what the results are meant to establish.

## 4.1 Simulation setup

The plant is the Eagar–Tsai solution for a moving Gaussian heat source on a semi-infinite domain (Eagar and Tsai, 1983), evaluated for the nickel-base alloy IN625. This solution is used because it is closed-form and reproduces the way the melt pool scales with power and speed. The point-source idealization overestimates the peak temperature, so the temperature field is used only to locate the melt pool boundary, which is taken as the liquidus isotherm at 1623 K. The width and depth are obtained from this boundary by stepping along the scan and taking the largest lateral and downward extents of the isotherm. The material is described by a density of 8440 kg $m^{-3}$, an absorptivity of 0.30, and temperature-dependent specific heat and conductivity. The model is calibrated against the melt pool dimensions reported for IN625 in the NIST AM-Bench benchmark AMB2018-02 (Lane et al., 2020). The power delivered in that benchmark differs from the commanded value, the commanded 150 W corresponding to 137.9 W delivered, and the calibration uses the delivered value. The model is anchored to this single benchmark; across the benchmark's power settings it reproduces the expected conduction-mode, monotonically ordered scaling of width and depth, but a broader calibration over several powers, speeds, and materials would require the instrumented measurements that Section 5 identifies as the principal validation step. A normalized enthalpy criterion is used to confirm that the operating range stays in the conduction regime and away from the onset of keyholing. The criterion is formed with the absorbed power and the material properties in the normalization for which the literature places the onset of keyholing near a value of 6, and the operating points used here fall between 2.6 and 3.8, which confirms conduction-mode melting.

Geometry is parameterized by the overhang angle $\theta$, measured so that $\theta$ = 90° is a flat, fully supported region and smaller angles are steeper overhangs. The reduced conduction under an overhang is represented by a confinement factor $C(\theta) = 1 + 0.7(1 - \theta/90)$. To reflect that the loss of conduction is mainly downward, this factor is applied to the depth as $C_z = C(\theta)$ and only weakly to the width as $C_y = 1 + 0.15(C(\theta) - 1)$. The directional form is a reduced-order assumption rather than a validated conduction model, and it is the simplest representation that produces a depth-to-width ratio which varies with geometry. With this calibration the ratio rises from 0.351 at $\theta$ = 90° to 0.462 at $\theta$ = 15°. The admissible depth is taken from a process-map form,

$$D_{lim}(\theta) = 25\ \mu m + 0.35 \mu m \cdot deg^{-1} \times \theta \tag{5}$$

which decreases as the overhang steepens, and a track is recorded as defective when its hidden depth exceeds $D_{lim}(\theta)$. These coefficients are chosen for the surrogate and would require calibration to a specific machine and material. The controller does not aim at this dross threshold directly but at an admissible depth set a fixed safety margin below it, 3 µm in what follows, and it is this admissible depth that is converted into the width bound of Section 3.4. The lack-of-fusion limit, which is a lower bound on depth, is maintained implicitly because the controller drives the width to the largest admissible value, and it is not scored separately in what follows.

Three materials give the Gaussian process a range of properties to learn from: IN625 and two variants with absorptivity 0.35 and 0.27, the latter also having a higher thermal diffusivity. The process is trained on single-track simulations of these materials at $\theta \in \{45°, 60°, 75°, 90°\}$, with the overhang angle and the material properties as inputs and the width gain, the width offset, and the ratio $\rho$ as outputs. A squared-exponential covariance is used, because these coefficients vary smoothly with angle and material, together with an additive noise term for the scatter in the training data. The lowest training angle, $\theta_{lo} = 45°$, sets the threshold below which the monotonic guard of Section 3.3 is active.

The knowledge layer is instantiated by importing the four reused ontologies and asserting the domain classes as subclasses of their classes, and by populating two calibration entities. The process is placed under bfo:process, the material and the melt pool under bfo:material entity, the laser power and scan speed under qudt:QuantityKind, the melt pool width and depth under sosa:ObservableProperty, the coaxial camera under sosa:Sensor, the width observation under sosa:Observation, and the defect modes under the geometric-flaw class of MASON, with lack of fusion and keyhole porosity as internal defects and dross and balling as surface defects of a small domain taxonomy that is itself placed under that class. After the imports the domain ontology resolves these parents to the external classes rather than to local placeholders. The instantiation comprises about 23 domain classes with seven object properties and 11 data properties. A geometry-calibration entity holds the coefficients of the admissible-depth relation, taken from process-map literature, and records that source as their provenance. A material-geometry-calibration entity holds the width gain, the width offset, and the ratio $\rho$ produced by the Gaussian process, with the standard deviation of the ratio, and a tag that identifies the inference as their source. The admissible-depth relation reads its coefficients from the first entity, and the width bound of Section 3.4 is formed from the admissible depth and the ratio held in the second, so that revising any coefficient is an edit to data.

The disturbances represent the dominant sources of run-to-run variation in LPBF. Absorptivity and delivered power are the inputs that most directly change the melt pool, and both vary during a build, the absorptivity with powder spreading, surface oxidation, and spatter, and the delivered power with the laser and the optical chain. Each is therefore perturbed multiplicatively. The variation is given a slow, temporally correlated component, modeled as a first-order autoregressive drift with correlation 0.85, and a fast, uncorrelated component, modeled as an independent random term. The slow drift is the disturbance the offset-free observer is intended to reject, and it is also the per-track analogue of the inter-layer accumulation studied in Section 4.2, so its inclusion tests both. The width measurement carries additive Gaussian noise with a standard deviation of 5 µm, so the controller acts on an imperfect width estimate. The magnitudes are kept to a few percent so that the geometric effect under study remains dominant and the comparison between controllers is not confounded by the disturbance. The depth is never measured and is used only to score defects, and all random streams are seeded so that runs are reproducible.

Each build is a sequence of 20 parallel tracks. A contiguous block of seven tracks is assigned to an overhang while the remaining tracks are flat, so each build contains an entry transition and an exit transition. Each track here stands for one hatch vector of the scan: the controller observes the width produced by a vector and applies the next command between vectors, the update rate to which the per-track timing budget of Section 4.3 refers. This within-layer action distinguishes the scheme from layer-to-layer depth control. The predictive controller uses a horizon of four scans, a working power range of 80 to 200 W, a rate bound of 60 W per track, and an observer gain of 0.5, with the move-suppression and slack weights of Section 3.4. The upper power is a qualified maximum for the material rather than the limit of the laser, so the flat region runs near it while the overhang requires a reduction that the rate bound spreads over more than one track. Four controllers are compared, with the names used throughout the paper. The geometry-blind controller applies the single flat-region ratio of 0.351 at every angle. The single-ratio controller applies one geometry-dependent ratio per angle, supplied by the knowledge layer; it is the controller used in the baseline of Section 4.2. The operating-power-aware controller, introduced in Section 4.3, refines it by evaluating that ratio at the commanded power. The oracle is given the true ratio of the plant and bounds the best result achievable by a ratio-based scheme. The reported metric is the number of defective tracks per build, averaged over the seeds, with the range across seeds reported alongside.

## 4.2 Geometry-conditioned defect prevention

These studies use the single geometry-dependent ratio and score the dross limit alone; Section 4.3 revisits the setting with the operating-power ratio and the lower, lack-of-fusion bound scored as well, and examines the robustness of the result. The first study uses a single overhang region and asks whether the knowledge layer prevents the dross that a geometry-blind controller produces. Table 1 reports the defective tracks for two overhang angles, averaged over six seeds. The geometry-blind controller left every overhang track defective at the steeper angle, whereas the single-ratio controller produced none in any seed at either angle.

**Table 1.** Defective tracks per build for a single overhang region (mean over six seeds).

| Overhang angle (depth limit) | Single-ratio | Geometry-blind |
|---|---|---|
| θ = 60° (46.0 µm) | 0.0 | 7.0 |
| θ = 45° (40.8 µm) | 0.0 | 6.2 |

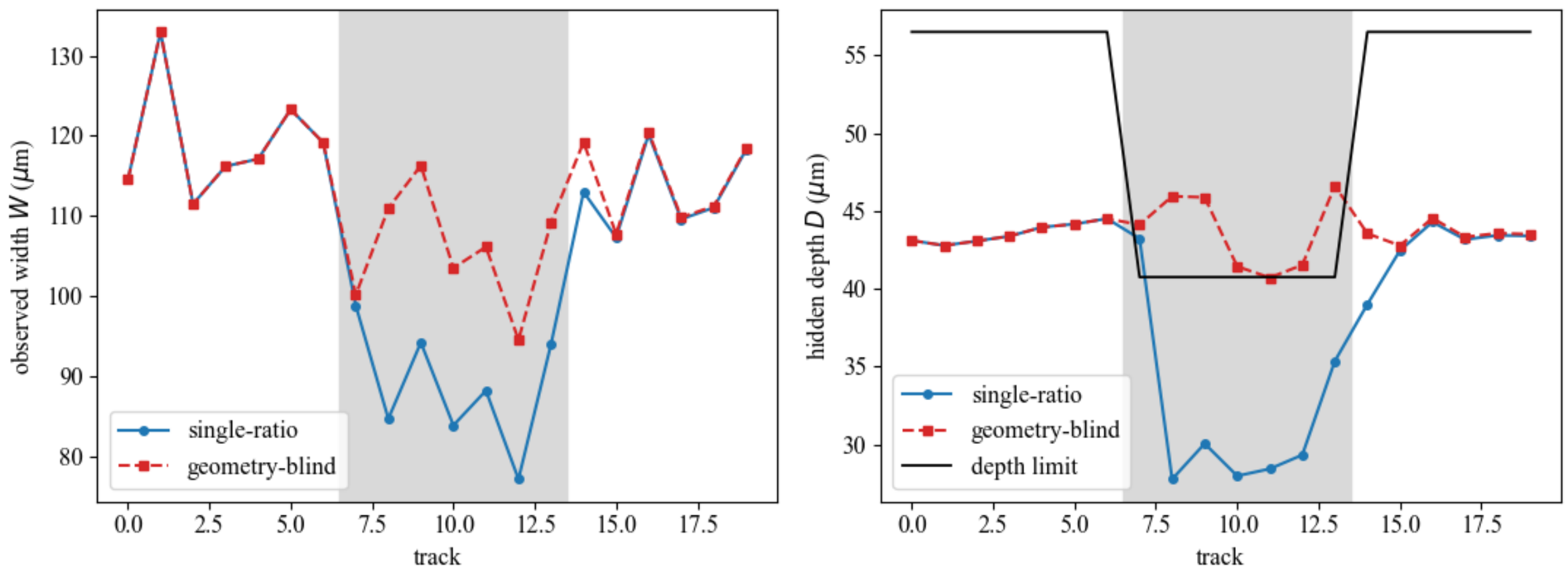

**Figure 3.** Width-only control across a θ = 45° overhang for one seed: observed width (left) and hidden depth with its limit (right).

Figure 3 shows the mechanism. The two controllers receive the same noisy width measurement. The geometry-blind controller applies the flat-region ratio, computes a width bound that is too high for the overhang, and allows a wider pool. The width it produces stays within the range it treats as safe, but the corresponding depth crosses the limit, as panel (b) shows. The single-ratio controller applies the larger ratio that holds in the overhang, sets a lower width bound, and keeps the depth below the limit throughout the region.

Two further results bear on the same comparison. When the geometric context was applied reactively, one track after a change in geometry rather than scheduled from the build plan, the controller produced two defective tracks at the overhang entry before recovering, because the command cannot fall far enough in a single track; scheduling the context along the path removed them. Section 4.3 examines this rate-bound trade-off in detail. Separately, when the controller read its coefficients entirely from the ontology, including the depth-limit coefficients and the learned ratio, it reproduced the result of the single-ratio controller, with no defects at $\theta$ = 45° against 6.2 for the geometry-blind controller. Editing a single stored coefficient confirmed that the control target follows the data: raising the slope of the admissible-depth relation from 0.35 to 0.40 changed the admissible depth at $\theta$ = 45° from 37.8 to 40.0 μm with no change to the control code. Holding the coefficients as data, rather than as constants inside the control code, therefore does not reduce control performance.

A controller that depends on calibrated coefficients must still behave safely below the calibrated range. With the Gaussian process trained on the angles in Section 4.1, the inferred ratio extrapolated to steeper angles with an error of a few thousandths, because it varies almost linearly with angle, and the monotonic guard changed it only slightly. This is interpolation of a smooth surrogate rather than a general guarantee; the guard biases the estimate to the safe side rather than adding accuracy.

**Table 2.** Defective tracks per build at θ = 40°, below the training range (mean over five seeds, range in parentheses).

| Controller | Defective tracks per build |
| --- | --- |
| Geometry-blind (flat-region ratio) | 4.2 (3–7) |
| Single-ratio ρ(θ) | 0.0 (0) |
| Single-ratio with guard | 0.0 (0) |
| Oracle (true ratio) | 0.0 (0) |

Table 2 reports a closed-loop build at θ = 40°, below the training range: the geometry-blind controller produced defects through most of the overhang, while the single-ratio controller and the oracle produced none, and the guard left this unchanged. Estimating the ratio from the observed width is also more robust to extrapolation than estimating depth directly, because the admissible depth is less regular in angle and reverts toward its mean when extrapolated, so the choice of width as the controlled variable carries a benefit beyond measurability.

The single-layer study does not account for the heat that builds up across layers, which raises the temperature of overhang regions as the part grows. To examine this effect, we added a reduced-order state that carries residual temperature from one layer to the next, following the macroscale treatment used in the heat-accumulation literature. With $T_{acc}$ the accumulated temperature, $\Delta t_d$ the inter-layer interval, $\tau$ a thermal time constant, $A$ the absorptivity, and $P_n$ the power of layer $n$, the state evolves as

$$T_{acc}[n+1] = e^{-\frac{\Delta t_d}{\tau}} T_{acc}[n] + \kappa \left(1 - \frac{1}{C(\theta)}\right) A P_n \tag{6}$$

The accumulation grows where downward conduction is poor, that is in overhangs, through the factor $1 - 1/C(\theta)$, and decays during the inter-layer interval. It raises the baseline temperature of the melt pool model, so the pool deepens over successive layers if the command is held fixed. The coefficients $\kappa$, $\tau$, and $\Delta t_d$ are chosen for the surrogate: the dwell and relaxation times are of the order of a layer interval for a small part, and the gain is fitted so that the accumulated temperature reaches a physically plausible steady level over the build (Table 4).

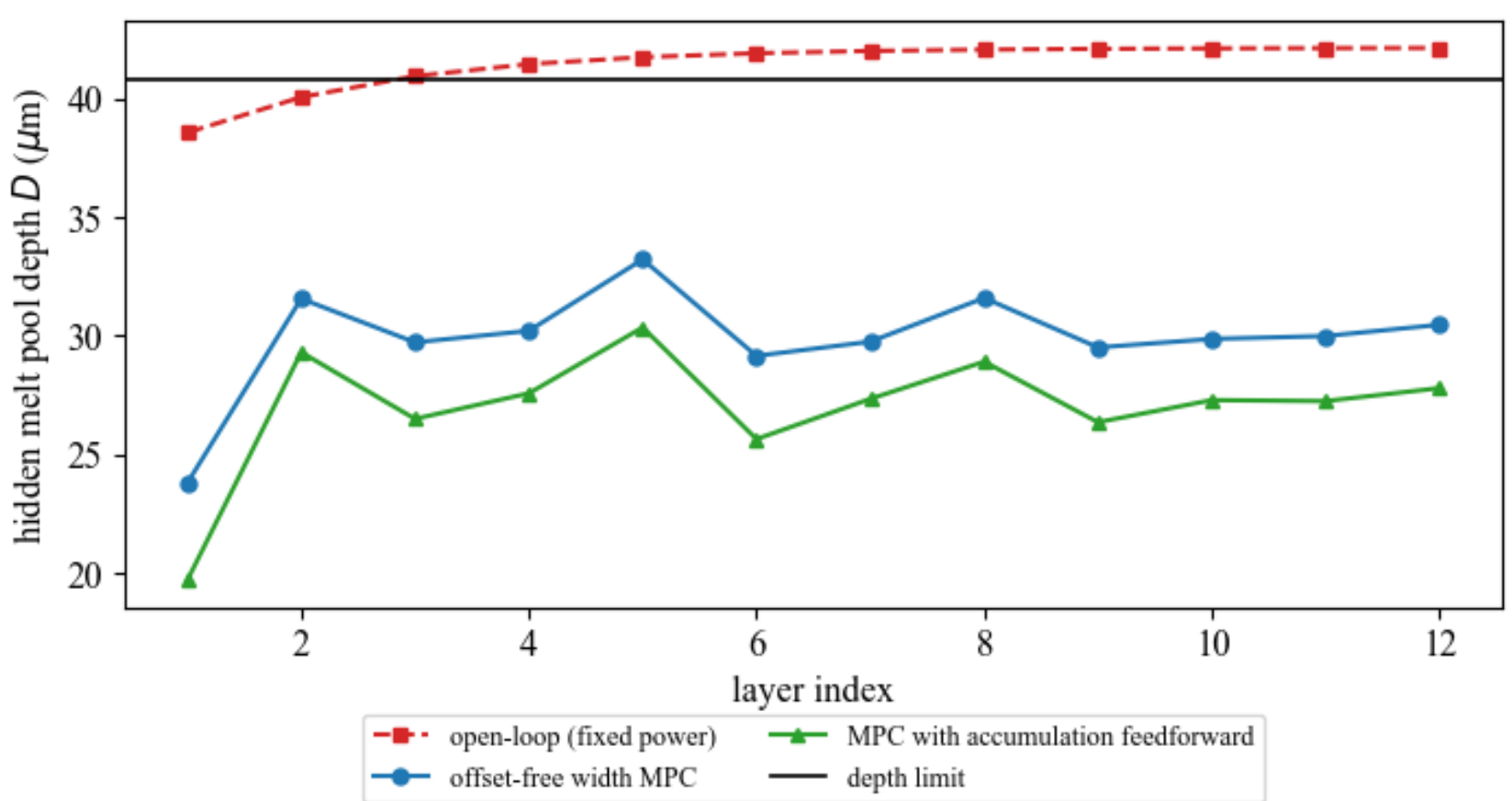

**Figure 4.** Hidden melt pool depth over 12 layers of a repeated θ = 45° overhang scan, for one seed.

**Table 3.** Defective layers and a thermal-gradient proxy over a 12-layer overhang build (mean over five seeds).

| Mode | Defective layers | Gradient proxy (mean) |
|---|---|---|
| Open-loop (fixed power) | 10.0 | 29.0 |
| Offset-free width MPC | 0.0 | 41.0 |
| With accumulation feedforward | 0.0 | 44.3 |

**Table 4.** Reduced-order inter-layer heat-accumulation coefficients used in Equation 6.

| Coefficient | Symbol | Value | Role |
|---|---|---|---|
| Inter-layer dwell | Δt_d | 10 s | time between successive layers at the same location (recoat interval) |
| Thermal relaxation time | τ | 18 s | decay time of the accumulated temperature between layers |
| Energy-to-temperature gain | κ | 6.5 $K \cdot W^{-1}$ (scaled) | maps absorbed power, reduced by downward confinement, to a temperature rise |

With a fixed command the depth drifted upward as heat accumulated and crossed the limit after the second layer, which gave 10 defective layers out of 12. The offset-free controller detected the rising width and reduced the power, and it held the depth below the limit for all 12 layers. Adding a feedforward term that anticipates the accumulation produced the same outcome with a smaller initial transient. This follows from the structure of the controller, since inter-layer accumulation appears as a slow drift and an offset-free controller removes slow drift without modification. The result is reported for five seeds, with no variation in the defect counts across seeds.

The same setup also bears on residual stress, which in LPBF is driven by the thermal gradient and the cooling rate near the melt pool, quantities the commanded power influences. Avoiding dross calls for lower power, which shrinks the pool and steepens the gradient, so the depth bound enforced here and a future bound on the gradient pull in opposite directions. Quantifying that trade-off requires a mechanical model and is left to future work; the gradient proxy reported in Table 3 is only a qualitative indicator of the trade-off, not a prediction of stress, and the residual-stress field itself is outside the scope of this study.

## 4.3 Power dependence, dual-mode scoring, and robustness

The depth-to-width ratio introduced in Section 3.2 was treated there mainly as a function of geometry. Examined across the operating range, it also depends on the laser power: between 80 and 200 W the ratio rises by 27 to 39 percent at every angle, because the depth and the width do not respond to power in the same proportion (Figure 5). The operating points stay in the conduction regime, with a normalized enthalpy below three, so this is an ordinary energy-density effect rather than a keyhole transition. A single value per angle is therefore representative only near the power at which it was measured, which is why the knowledge layer supplies the ratio at the power the controller is about to command (Section 3.2).

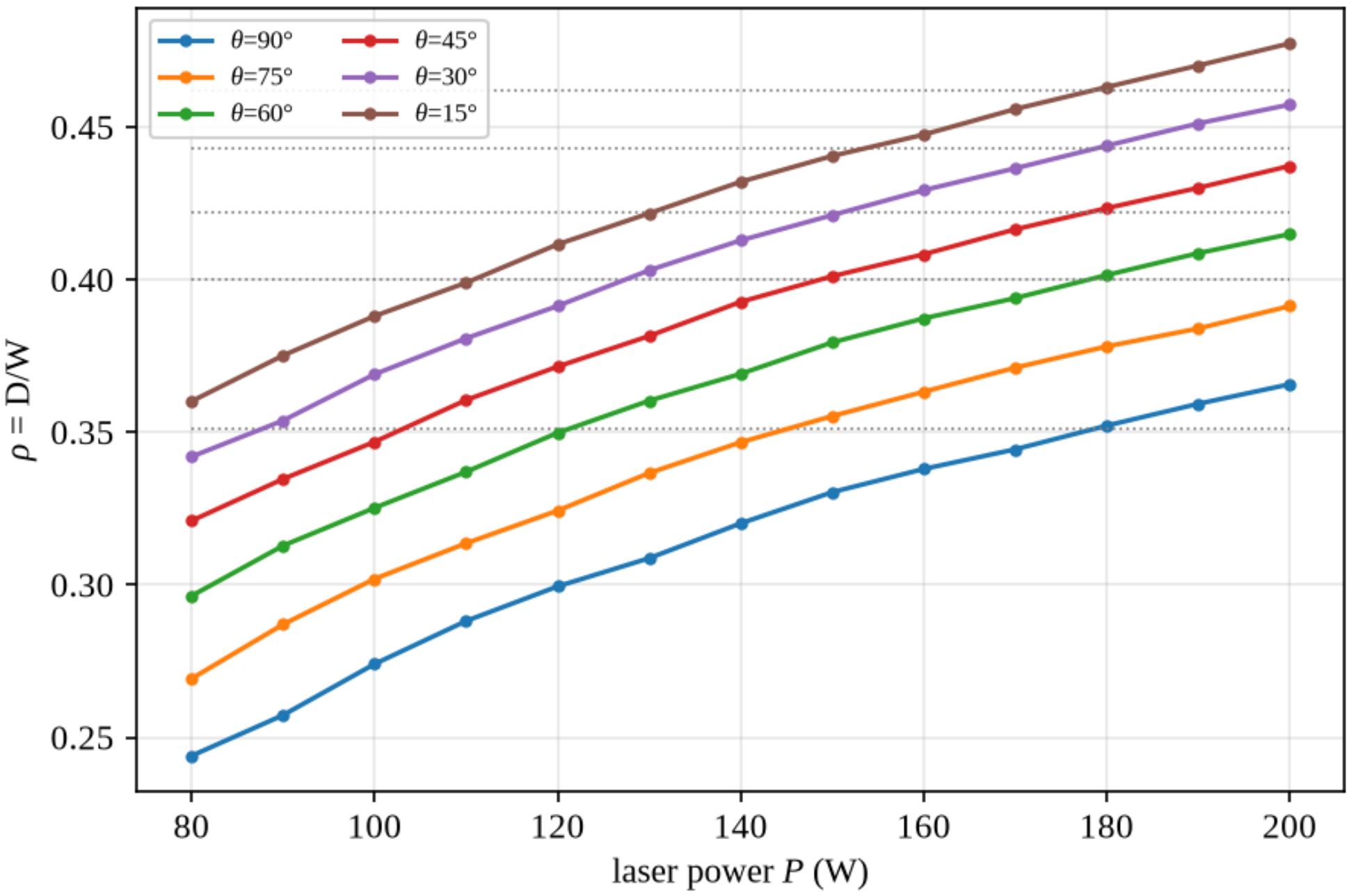

**Figure 5.** Depth-to-width ratio ρ = D/W against laser power for each overhang angle; dotted lines mark the single value used in the baseline.

Scoring only the upper failure mode can conceal the lower one. When the admissible depth is bounded below by a lack-of-fusion floor as well as above by the dross limit, a controller that uses a single ratio calibrated at high power drives the overhang pool, where it has reduced the power, to the shallow edge of the band and across the fusion floor, even though no dross forms (Figure 6). The earlier report of zero defective tracks reflected scoring the dross limit alone; once both bounds are scored, the single-ratio controller trades dross for occasional lack of fusion, while the geometry-blind controller shows the opposite failure, crossing the dross limit in the overhang.

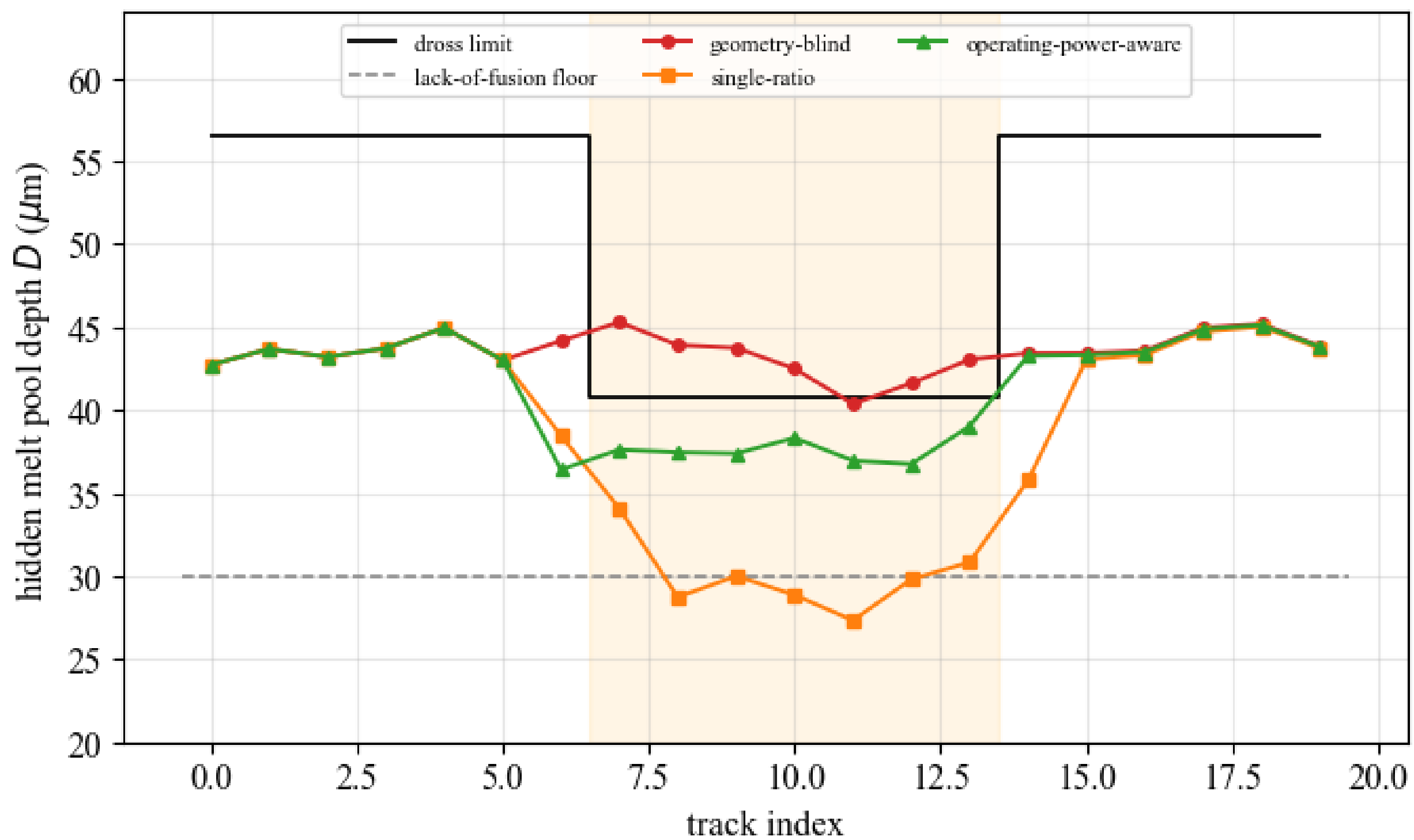

**Figure 6.** Hidden depth along a build with a θ = 45° overhang (shaded), for three controllers.

Two changes remove both modes. Evaluating the ratio at the operating power, with a short fixed-point pass so that the assumed ratio and the commanded power are mutually consistent, centers the pool in the band rather than leaving it at the shallow edge. Sizing the safety margin to the centre of the admissible band, rather than three micrometres below its ceiling, then keeps headroom on both sides: a margin sweep shows that values near the band centre, about seven micrometres for the angle studied, give zero dross and zero lack of fusion across seeds, whereas the thinner three-micrometre margin leaves a few dross tracks once the power dependence is taken into account.

The first concern is whether the result is an artifact of a surrogate that the controller has itself calibrated. To break that circularity the ratio was calibrated on one plant and the closed loop was run on deliberately mismatched plants whose downward heat loss was raised by eight percent or made to depend on power, changes the controller's calibration does not contain. Table 5 reports the outcome. The geometry-blind controller produces dross in almost every overhang track regardless of the mismatch; the single-ratio controller removes the dross but, using a ratio fixed at high power, under-fuses and leaves several lack-of-fusion tracks; the operating-power-aware controller, with its margin sized to the admissible band, holds dross at zero across all of the mismatched plants and leaves only one to two residual lack-of-fusion tracks, the same enforcement-accuracy limit noted earlier. A systematic mismatch shifts the safe margin window rather than defeating the controller: under the eight-percent sink the dross-free margin moves from about six to about nine micrometres. The disturbance magnitude was also raised severalfold, which degraded the result gradually, adding at most one or two dross tracks rather than failing abruptly.

**Table 5.** Defective tracks per build (dross / lack-of-fusion, mean over four seeds) on deliberately mismatched plants.

| Plant relative to controller belief | geometry-blind | single-ratio ρ(θ) | operating-power-aware ρ(θ, P) |
|---|---|---|---|
| no mismatch | 6.0 / 0 | 0 / 4.2 | 0 / 1.2 |
| downward sink +8% | 7.0 / 0 | 0 / 2.2 | 0 / 0.5 |
| power-dependent depth | 5.8 / 0 | 0 / 5.5 | 0 / 2.0 |
| deeper sink + power-dependent | 6.8 / 0 | 0 / 4.0 | 0 / 1.2 |

The per-scan optimization is a small quadratic program, and its cost must fit between tracks. Table 6 separates the solve time from the one-time problem construction. When the compiled problem is reused, the solve takes about two milliseconds at the ninety-fifth percentile, roughly one-sixth of the time available between tracks at a typical scan speed, so the per-track update is feasible in principle; the larger figure sometimes quoted belongs to rebuilding the problem in the prototype and does not lie on the per-track path.

**Table 6.** Per-scan optimization timing against the available track period.

| Quantity | Value |
|---|---|
| QP solve time, mean (reused compiled problem) | 1.56 ms |
| QP solve time, 95th percentile | 1.93 ms |
| QP solve time, maximum | 3.18 ms |
| Track period (9 mm hatch at 0.8 m/s) | 11.2 ms |
| 95th-percentile solve as fraction of track period | 17% |
| Problem construction in the prototype (not on the per-track path) | ~21 ms |

A separate trade-off governs the transitions into and out of an overhang. When the geometric context is supplied only after a change of geometry, the rate bound on the command limits how fast the power can fall, and a few tracks at the entry exceed the dross limit before the controller catches up. Figure 7 sweeps the rate bound: loosening it reduces the transition defects but makes the commanded power jump more sharply, while tightening it smooths the command at the cost of more transition defects. Scheduling the context along the known scan path removes the transition defects altogether at a physically realistic rate bound, which is why the architecture schedules rather than reacts.

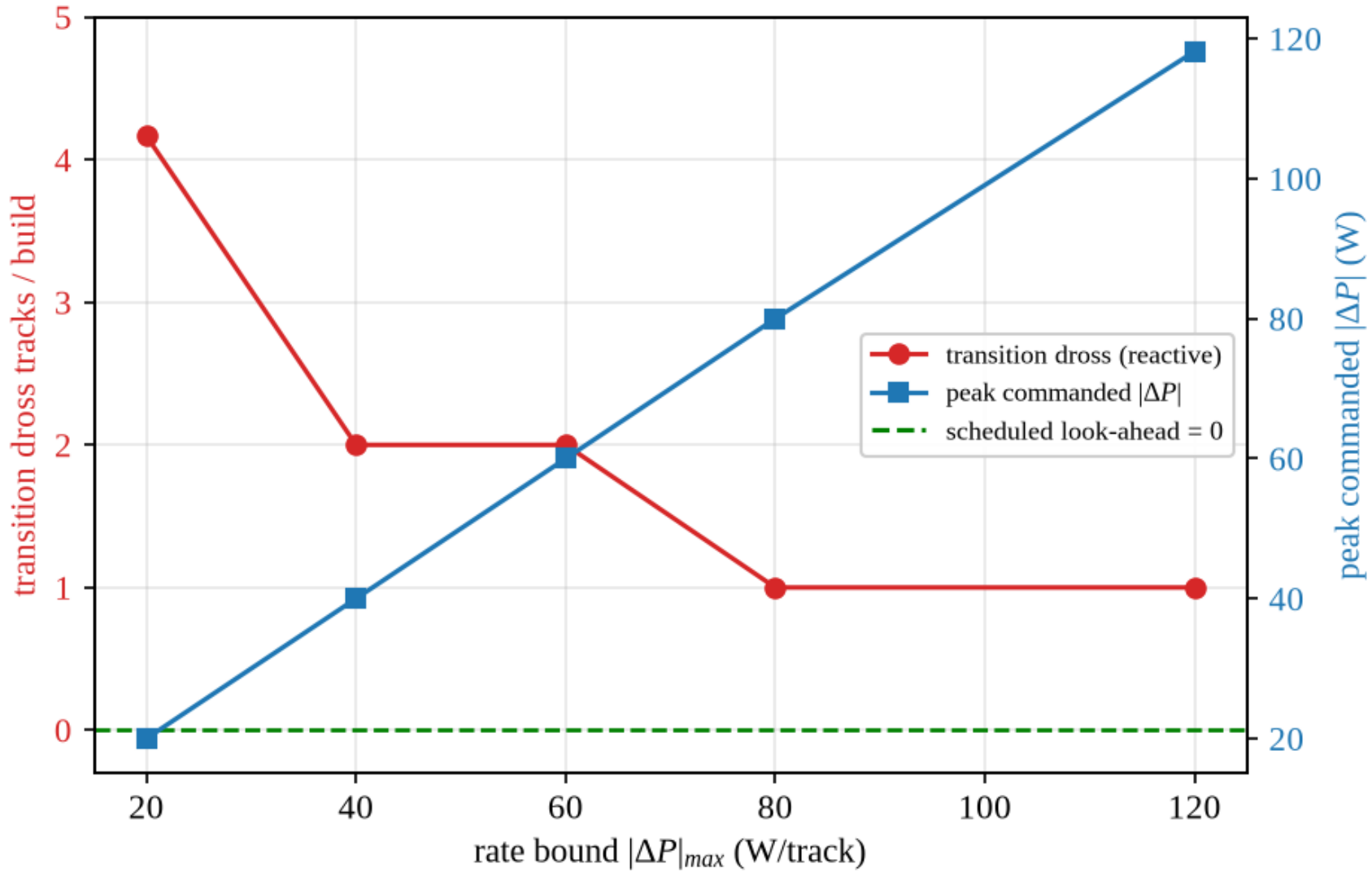

**Figure 7.** Transition defects and command smoothness versus the rate bound, with the geometric context applied reactively.

Finally, the control result does not depend on the particular confinement coefficients chosen for the surrogate. Varying them over a wide range shifts the numerical ratios but not the outcome, because the controller re-measures the ratio from whatever plant it faces, whereas the geometry-blind controller fails throughout. Together these checks indicate that the defect prevention is a property of the architecture rather than of one self-consistent parameter set, though they do not replace the experimental validation identified in Section 5 as the principal next step.

## 4.4 Reasoning, extensibility, and material retargeting of the knowledge layer

Several properties of the knowledge layer were examined apart from the control results, because they bear on what the ontology contributes beyond a typed configuration file.

One property is the reuse of external knowledge. The four community ontologies named in Section 3.3 were imported, and after the import the domain classes resolved to the external classes rather than to local placeholders, so the process, the parameters, the melt pool observables, and the defect modes each became a subclass of the corresponding imported class. The control results are unchanged by the import, which affects the description of the model and its interoperability rather than its numerical behavior.

The qualitative part of the layer was expressed as logical class definitions: a feature whose overhang angle does not exceed a threshold is dross-prone, one below the calibrated range is additionally out of calibration, and one whose normalized enthalpy exceeds a declared value is keyhole-prone. A description-logic reasoner checked the resulting ontology and found no unsatisfiable class, so the definitions are mutually consistent. The same reasoner answered a set of competency questions by inference rather than by assertion: it classified a thirty-degree feature as dross-prone and a seventy-five-degree feature as not, classified a forty-degree feature as both dross-prone and out of calibration, and derived that every out-of-calibration feature is necessarily dross-prone, a subsumption that was never stated directly. The qualitative classification used to select a control mode is therefore obtained from the definitions, not hand-coded.

The constraint sets used above are not fixed in the controller; they are decided in the loop by the reasoner. At each context the upcoming feature is classified by a description-logic reasoner from its overhang angle and energy density, and the inferred classes determine which bounds are active: the dross cap in every context, the lack-of-fusion floor in overhangs, the monotone guard below the calibrated range, and, where a process window is declared, an energy-density cap (Figure 8). Because the classification depends only on the context, the reasoner is invoked a few times per build, three times for the case shown, and its result is cached, so the per-scan path remains the single quadratic program timed above; each reasoning call took on the order of a second with a standard engine, off the per-track path. Deciding the active set by inference rather than by a fixed rule is what makes the layer extensible and selective, as developed below. What the inference does not by itself guarantee is that an activated bound is met exactly; enforcing the fusion floor in particular is only as accurate as the depth-to-width ratio at the operating power and the controller's behavior through the geometry transition, which is the limit noted earlier.

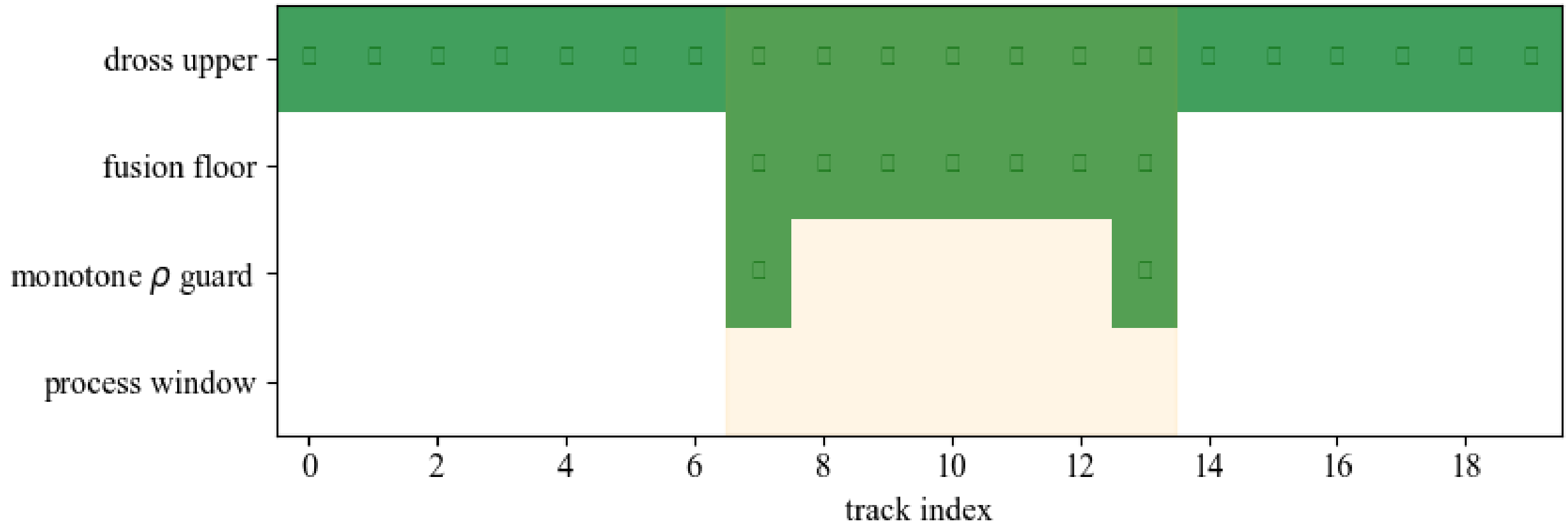

**Figure 8.** Active constraint set inferred by the in-loop reasoner along a build with a θ = 45° overhang (edges at θ = 40°).

A further property is extensibility: because the controller enforces every constraint declared by the knowledge layer, a constraint that was not part of the original model can be added as data. A process-window limit on the normalized enthalpy, chosen to keep melting in the conduction regime, was added as a single entity in the ontology with an upper value of 2.5. With this constraint present the controller lowered the flat-region power from about 199 to about 184 W, which brought the normalized enthalpy under the chosen value, and it did so with no defective tracks and no change to the control law. The depth bound continued to govern the overhang, where it remained the more restrictive of the two. This indicates that the relational forms of the model are fixed while the set of active constraints and the choice of material are open to extension through data.

The same separation of form from data lets the controller be retargeted to a new material by replacing data alone. Substituting nominal high-temperature properties for 316L stainless steel and for Ti-6Al-4V, which differ from the nickel alloy in density, diffusivity, liquidus, and absorptivity rather than in absorptivity alone, the statistical component re-estimates the width model and the ratio, and the controller prevents dross and avoids lack of fusion alike on each, scored as in Section 4.3, with no change to the control law.

These properties are what distinguish the knowledge layer from a lookup table with conditional rules. The import of community vocabularies makes the model interoperable rather than tied to private fields. The reasoner selects the active set of constraints for each context, which a fixed rule set cannot do without enumerating the contexts by hand; applying the union of all constraints everywhere instead would cap the flat-region power needlessly, as the energy-density limit does when it is left active outside the regions that need it. And a new constraint or material enters as data, after which the controller enforces it without any change to its code. A typed configuration file provides none of these on its own.

These examples each exercise a different axis of the architecture rather than repeating one result. The depth-from-width mapping expresses an objective stated on an unobservable variable as a bound on an observable one; the energy-density cap adds a new constraint as data; the change of material loads a new operating context as data; and the per-context activation has the reasoner choose among declared constraints by inference. The generality claimed for the architecture is the span of these demonstrated axes within laser powder bed fusion, not an extrapolation from a single case.

## 5. Conclusion

This paper set out to address a gap between two established approaches to melt pool control in laser powder bed fusion: closed-loop controllers that reject process disturbances but are blind to geometry, and feedforward schemes that anticipate geometry but cannot react to a live signal. We proposed a geometry-conditioned, neuro-symbolic closed-loop architecture in which an ontology is placed inside the control loop and converts the geometry and material of each upcoming scan into the references and constraints of a constraint-aware predictive controller. In this architecture the ontology connects the control objectives and constraints of the process to the signals the controller can observe, and a reasoner maintains that connection as the geometry changes, so that a bound on a quantity that cannot be sensed directly becomes a bound on one that can. The case study evaluated here is overhang dross, in which the unobservable objective is the melt pool depth and the observable signal is the width: the ontology supplies a geometry-dependent ratio that turns a limit on the hidden depth into a limit on the observed width, and the controller regulates the width it can see while respecting the depth limit it cannot.

Evaluated in an Eagar–Tsai surrogate calibrated to the NIST AM-Bench melt pool benchmark for IN625, the architecture removed the overhang dross that a geometry-blind controller produced. Under the dual scoring that Section 4.3 shows is necessary, the operating-power-aware controller holds dross at zero across the angles and mismatches tested while leaving a small residual lack of fusion of one to two tracks, set by the accuracy of the ratio at the operating power; the single-mode figure in which all seven overhang tracks become defect-free describes only the dross limit. The ratio inferred by the Gaussian process extrapolated safely below the calibrated range, an interpolation of a smooth surrogate rather than a guarantee, with a monotonic guard biasing the estimate toward the safe side. When heat accumulation across layers was modeled, the offset-free controller absorbed the resulting slow drift and held the depth below its limit over a 12-layer build that drifted into dross under a fixed command.

Several properties of the knowledge layer were examined apart from the control results. The four community ontologies it builds on were imported rather than imitated, so that the process, the quantities, the sensing concepts, and the defect modes resolve to established external classes. The calibration coefficients were held as data with their provenance, which let a new process-window constraint and additional materials be introduced without any change to the control law, and editing a single stored coefficient shifted the control target as expected. The qualitative classification that selects a control mode was shown to be derived in the loop by a description-logic reasoner, which selects the active constraints while the set-point is computed arithmetically. Together these indicate that the symbolic layer is not a fixed rule base but an interoperable and extensible model.

The study has clear limits. The relational forms at the core of the model, the linear admissible-depth law, the single depth-to-width ratio, and the directional confinement factor, are calibrated to the surrogate and would need re-examination and experimental calibration for a specific machine, material, and process regime. The plant is a closed-form thermal solution rather than a high-fidelity melt pool simulation, the depth used to score defects is never measured but computed from that plant, and the residual-stress discussion rests on a thermal-gradient proxy rather than a mechanical model. The reasoner now runs inside the loop, selecting the active set of constraints on each change of context while the set-point is evaluated arithmetically for speed; enforcing an activated bound exactly, however, is only as good as the ratio at the operating point and the controller's behavior through a geometry transition, which bounds how completely the lower, lack-of-fusion bound is met. A closer analysis in Section 4.3 further shows that the ratio varies with the operating power and that the lower, lack-of-fusion bound must be scored alongside the dross limit; the operating-power form of the ratio and a band-centred margin are introduced there to keep both bounds satisfied. The mismatch used to probe circularity stays within the Eagar–Tsai family, a parameter perturbation rather than an independent physical model, so a cross-model check against other simulation or experimental data is needed to test the ratio more stringently. The timing headroom is likewise reported at one favorable operating point, a 9 mm hatch at 0.8 m per second; at faster speeds or shorter vectors, where the time between tracks approaches the solve time, the per-track guarantee narrows, and the applicable envelope is part of what hardware testing must establish. For these reasons the results establish the feasibility of placing geometric knowledge inside the LPBF control loop, not the performance of a controller deployed on hardware.

Future work follows directly from these limits. The most immediate step is calibration and validation on an instrumented machine with coaxial melt pool sensing, which would test whether the width-to-depth relationship supplied by the ontology holds in practice. Beyond that, the knowledge layer invites extension: additional in-process constraints such as a bound on the thermal gradient can be declared as data and enforced alongside the depth bound, the schema can be broadened past the dross-centric forms used here to other geometry-induced defects, and the reasoner, already used in the loop to maintain the active set of constraints as the build proceeds, can be extended to richer defect taxonomies. The broader contribution is the demonstration that an explicit, standards-aligned knowledge layer can occupy the place inside a process-control loop that is usually reserved for fixed gains or opaque policies, with overhang dross control as the instance demonstrated here, and can do so without sacrificing the real-time behavior that control requires.

**CRediT authorship contribution statement**

Gisuk Hong: Conceptualization, Methodology, Software, Validation, Formal analysis, Investigation, Writing – original draft, Visualization.

Jaebong Cho: Software, Validation, Investigation, Writing – review & editing.

Hyunbo Cho: Conceptualization, Supervision, Project administration, Resources, Writing – review & editing.

**Statements and Declarations**

**Competing interests**

The authors declare that they have no known competing financial interests or personal relationships that could have appeared to influence the work reported in this paper.

**Funding**

The authors did not receive any specific grant from funding agencies in the public, commercial, or not-for-profit sectors.

**Ethics approval**

Not applicable. This study did not involve human participants or animals.

**Consent to participate and consent for publication**

Not applicable.

**Data availability**

The code, the calibrated surrogate, the ontology, and the scripts that generate the reported figures and tables are available from the corresponding author upon reasonable request. No proprietary or experimental data were used; the surrogate is calibrated to the publicly available NIST AM-Bench AMB2018-02 benchmark.